\documentclass[10pt,twocolumn,letterpaper]{article}

\usepackage[pagenumbers]{cvpr} % To force page numbers, e.g. for an arXiv version

\usepackage{multirow}
\usepackage{graphicx}
\usepackage{amsmath}
\usepackage{amssymb}
\usepackage{booktabs}
\usepackage[accsupp]{axessibility}

\usepackage[pagebackref,breaklinks,colorlinks]{hyperref}

\usepackage[capitalize]{cleveref}
\crefname{section}{Sec.}{Secs.}
\Crefname{section}{Section}{Sections}
\Crefname{table}{Table}{Tables}
\crefname{table}{Tab.}{Tabs.}

\def\cvprPaperID{7794} % *** Enter the CVPR Paper ID here
\def\confName{CVPR}
\def\confYear{2023}

\begin{document}

%%%%%%%%% TITLE - PLEASE UPDATE
\title{Class Attention Transfer Based Knowledge Distillation}

\author{Ziyao Guo\textsuperscript{1}, Haonan Yan\textsuperscript{1,2,}\thanks{Corresponding author} , Hui Li\textsuperscript{1,}\footnotemark[1] , Xiaodong Lin\textsuperscript{2}\\
\textsuperscript{1}Xidian University, \textsuperscript{2}University of Guelph\\
{\tt\small gzyaftermath@outlook.com}
% For a paper whose authors are all at the same institution,
% omit the following lines up until the closing ``}''.
% Additional authors and addresses can be added with ``\and'',
% just like the second author.
% To save space, use either the email address or home page, not both
%
% \and
% Second Author\\
% Institution2\\
% First line of institution2 address\\
% {\tt\small secondauthor@i2.org}
}
\maketitle
\begin{abstract}
Previous knowledge distillation methods have shown their impressive performance on model compression tasks, however, it is hard to explain how the knowledge they transferred helps to improve the performance of the student network. In this work, we focus on proposing a knowledge distillation method that has both high interpretability and competitive performance. We first revisit the structure of mainstream CNN models and reveal that possessing the capacity of identifying class discriminative regions of input is critical for CNN to perform classification. Furthermore, we demonstrate that this capacity can be obtained and enhanced by transferring class activation maps. Based on our findings, we propose class attention transfer based knowledge distillation (CAT-KD). Different from previous KD methods, we explore and present several properties of the knowledge transferred by our method, which not only improve the interpretability of CAT-KD but also contribute to a better understanding of CNN. While having high interpretability, CAT-KD achieves state-of-the-art performance on multiple benchmarks. Code is available at: \url{https://github.com/GzyAftermath/CAT-KD}.
\end{abstract}

%-------------------------------------------------------------------------
\begin{figure}
\centering
\includegraphics[width=8cm]{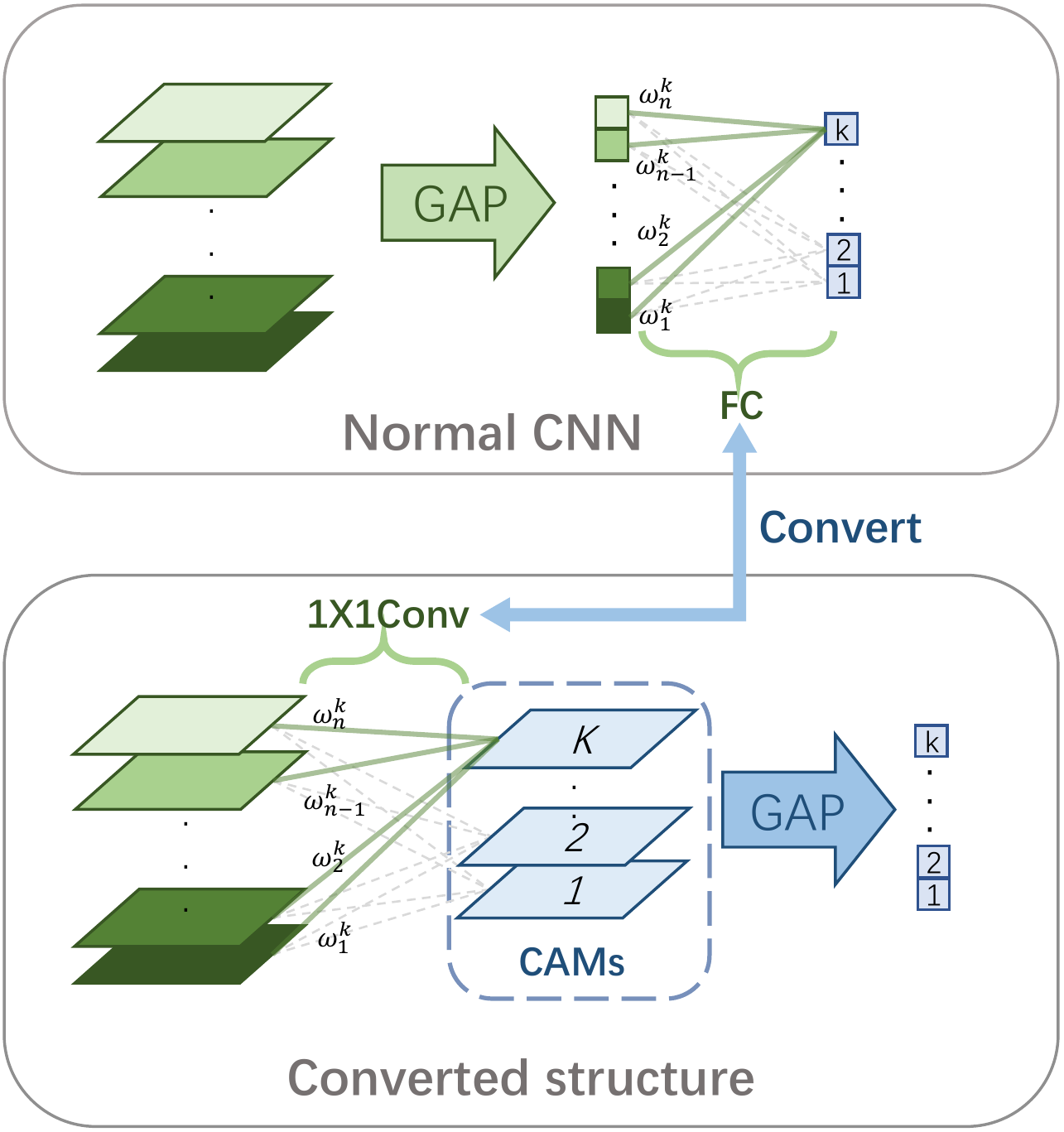}
\caption{Illustration of the converted structure. After converting the FC layer into a convolutional layer with 1$\times$1 kernel and moving the position of the global average pooling layer, CAMs can be obtained during the forward propagation.} 
\label{fig:main}
\end{figure}

%%%%%%%%% BODY TEXT
\section{Introduction}
\label{sec:intro}
\begin{figure*}
\centering
\includegraphics[width=3.4cm]{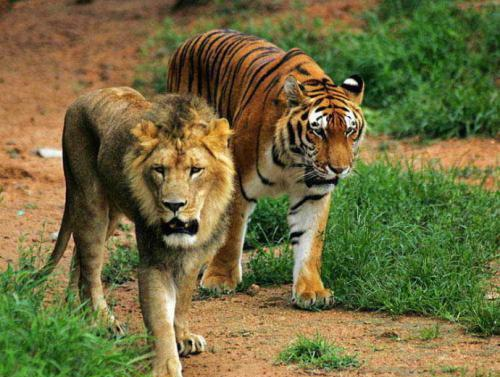}
\hfill
\includegraphics[width=3.4cm]{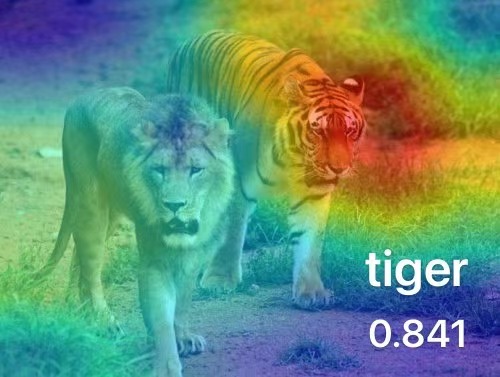}
\hfill
\includegraphics[width=3.4cm]{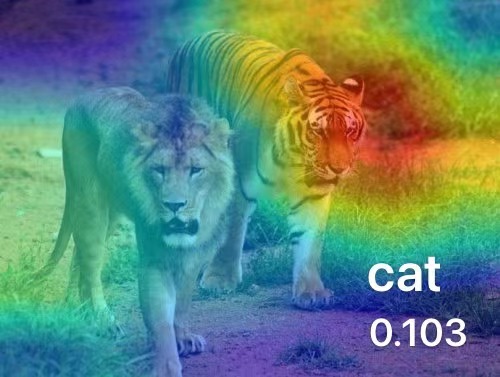}
\hfill
\includegraphics[width=3.4cm]{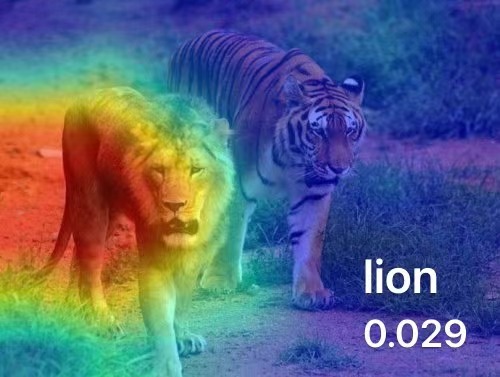}
\hfill
\includegraphics[width=3.4cm]{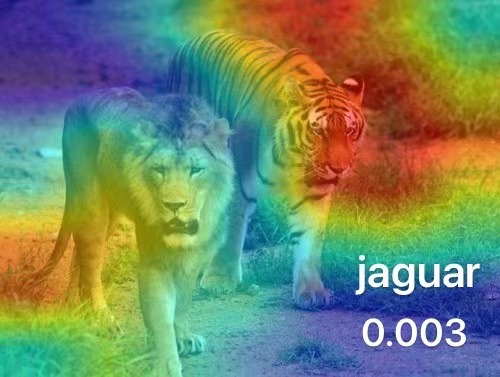}
\caption{Visualization of CAMs corresponding to categories with Top 4 prediction scores for the given image. The predicted categories and their scores are reported in the picture.}
\label{fig:CAMs}
\end{figure*}
Knowledge distillation (KD) transfers knowledge distilled from the bigger teacher network to the smaller student network, aiming to improve the performance of the student network. Depending on the type of the transferred knowledge, previous KD methods can be divided into three categories: based on transferring logits \cite{GeoffreyEHinton2015DistillingTK,KD2,KD3,KD4,DKD}, features \cite{FitNets,RKD,OFD,CRD, reviewKD, feature1, feature2, feature3}, and attention \cite{AT}. Although KD methods that are based on transferring logits and features have shown their promising performance \cite{DKD,reviewKD}, it is hard to explain how the knowledge they transferred helps to improve the performance of the student network, due to the uninterpretability of logits and features. Relatively, the principle of attention-based KD methods is more intuitive: it aims at telling the student network which part of the input should it focus on during the classification, which is realized by forcing the student network to mimic the transferred attention maps during training. However, though previous work AT \cite{AT} has validated the effectiveness of transferring attention, it does not present what role attention plays during the classification. This makes it hard to explain why telling the trained model where should it focus could improve its performance on the classification mission. Besides, the performance of the previous attention-based KD method \cite{AT} is less competitive compared with the methods that are based on transferring logits and features \cite{DKD, reviewKD}. In this work, we focus on proposing an attention-based KD method that has higher interpretability and better performance.
We start our work by exploring what role attention plays during classification. After revisiting the structure of the mainstream models, we find that with a little conversion (illustrated in \Cref{fig:main}), class activation map (CAM) \cite{CAM}, a kind of class attention map which indicates the discriminative regions of input for a specific category, can be obtained during the classification. Without changing the parameters and outputs, the classification process of the converted model can be viewed in two steps: (1) the model exploits its capacity to identify class discriminative regions of input and generate CAM for each category contained in the classification mission, (2) the model outputs the prediction score of each category by computing the average activation of the corresponding CAM. Considering that the converted model makes predictions by simply comparing the average activation of CAMs, possessing the capacity to identify class discriminative regions of input is critical for CNN to perform classification. The question is: can we enhance this capacity by offering hints about class discriminative regions of input during training? To answer this question, we propose class attention transfer (CAT).

During CAT, the trained model is not required to predict the category of input, it is only forced to mimic the transferred CAMs, which are normalized to ensure they only contain hints about class discriminative regions of input. Through experiments with CAT, we reveal that transferring only CAMs can train a model with high accuracy on the classification task, reflecting the trained model obtains the capacity to identify class discriminative regions of input. Besides, the performance of the trained model is influenced by the accuracy of the model offering the transferred CAMs. This further demonstrates that the capacity of identifying class discriminative regions can be enhanced by transferring more \textit{precise} CAMs.

Based on our findings, we propose class attention transfer based knowledge distillation (CAT-KD), aiming to enable the student network to achieve better performance by improving its capacity of identifying class discriminative regions. Different from previous KD methods transferring \textit{dark knowledge}, we present why transferring CAMs to the trained model can improve its performance on the classification task. Moreover, through experiments with CAT, we reveal several interesting properties of transferring CAMs, which not only help to improve the performance and interpretability of CAT-KD but also contribute to a better understanding of CNN. While having high interpretability, CAT-KD achieves state-of-the-art performance on multiple benchmarks. Overall, the main contributions of our work are shown below:
\begin{itemize}
\vspace{-4pt}
\item We propose class attention transfer and use it to demonstrate that the capacity of identifying class discriminative regions of input, which is critical for CNN to perform classification, can be obtained and enhanced by transferring CAMs. 
\item We present several interesting properties of transferring CAMs, which contribute to a better understanding of CNN.
\item We apply CAT to knowledge distillation and name it CAT-KD. While having high interpretability, CAT-KD achieves state-of-the-art performance on multiple benchmarks.
\end{itemize}

%-------------------------------------------------------------------------

%------------------------------------------------------------------------

\section{Background}
\label{sec:formatting}

%-------------------------------------------------------------------------

The concept of knowledge distillation was proposed in \cite{GeoffreyEHinton2015DistillingTK}. As a transfer learning method, KD aims to improve the performance of the smaller student network by transferring the \textit{dark knowledge} distilled from the bigger teacher network. Previous KD methods can be divided into three types: distillation from logits \cite{GeoffreyEHinton2015DistillingTK,KD2,KD3,KD4,DKD}, features \cite{FitNets,RKD,OFD,CRD, reviewKD, feature1, feature2, feature3} and attention \cite{AT}.

To our knowledge, AT \cite{AT} is the only KD method based on transferring attention, which defines attention map as the spatial map indicating the area of input that the model focus on most. In practice, they obtain attention maps by calculating the sum of feature maps while their values are absolutized. However, AT did not present what role \textit{attention} plays during the classification and why transferring attention maps defined in this way can improve the performance of the student network.
%Apart from directly transferring attention, a few KD methods utilize attention mechanisms to transfer feature maps. \cite{attention_feature_mapping} utilized the attention mechanism to identify similar features between the teacher and student, the resulting similarities are then used to control the distillation intensities for all possible feature pairs. \cite{reviewKD} proposed an attention-based fusion module to aggregate feature maps that come from different levels.
% However, although the high-level attention map generated in this way indicates the discriminative regions of the target category, considering that high-level feature maps contain information related to other categories \cite{CAM} and the feature maps generated by different models may have different numbers of channels, it is unsuitable to define attention map in this way. However, although the high-level attention map generated in this way indicates the discriminative regions of the target category, AT also requires intermediate feature maps that have little connection with attention, making it not entirely based on attention transfer. Besides, the effect of AT is less competitive compared with previous feature-based state-of-the-art KD methods.

Previous works related to class attention originate from \cite{CAM}, where the authors propose to utilize high-level feature maps and the parameters of the fully connected layer to generate attention map for a specific category, which is named class activation map (CAM). According to \cite{CAM}, class discriminative regions of input are highlighted in the corresponding CAM. To facilitate understanding, we visualize several CAMs in \Cref{fig:CAMs}. The following works have successfully applied CAM in various weakly supervised visual tasks \cite{CAMap1,CAMap2,CAMap3}. Besides, there are also many works focus on generalizing CAM \cite{GradCAM, GradCAM++,Score-CAM} and improving the performance of models by exploiting the information contained in CAM during training \cite{CAM_in_training1,CAM_in_training2}. 

Previous works have not presented what role attention plays during classification and why transferring attention maps can improve the trained model's performance on the classification mission. In this paper, we focus on figuring out this question and try to propose an attention-based KD method that has both high interpretability and competitive performance.

%------------------------------------------------------------------------

\section{Our Method}

In this section, we first analyze the structure of the mainstream CNN models and reveal that possessing the capacity of identifying class discriminative regions is critical for CNN to perform classification. Then we further propose class attention transfer to prove that this capacity can be obtained and enhanced by transferring CAMs. Finally, we apply CAT to knowledge distillation.

%-------------------------------------------------------------------------

% However, although the features after global average pooling (GAP) contain enough information for classification, this process results in a loss of information. Through the exploration of the feature maps generated before GAP, Zhou \textit{et al.} show that class activation map (CAM), which indicates the discriminative regions corresponding to a specific category of the input image, can be obtained by computing the weighted sum of the feature maps generated by the last convolutional layer \cite{CAM}. 

% However, with a little tweaking, CAM is an intermediate product of the classification process.

\subsection{Revisit the structure of CNN}
\label{sec: structure}

In image classification tasks, mainstream models usually use CNN to extract features, the resulting high-level feature maps are then globally pooled and fed to a simple fully connected layer to perform classification \cite{densenet, resnet, pyramidnet}. Let $\mathbf{F}=[F_{1},F_{2},...,F_{C}]\in\mathbb{R}^{C\times W \times H}$ represents the feature maps generated by the last convolutional layer, where $C$, $W$, and $H$ indicate channel dimension, width, and height respectively. And $f_{j}(x,y)$ denotes the activation of $\mathbf{F}$ in $j$ channel at spatial location $(x,y)$, while $\mathit{GAP}$ is the global average pooling layer. Then the process of calculating logits for normal CNN models can be written as:
\begin{equation}
\label{eq_logit_Normal}
\begin{aligned}
L_{i}&=\sum_{1\leq j \leq C} \omega _{j} ^{i} \times \mathit{GAP}(F_{j})\\
&=\dfrac{1}{W \times H}\sum_{x,y} \sum_{1\leq j \leq C} \omega _{j} ^{i} \times f_{j}(x,y),
\end{aligned}
\end{equation}
where $L_{i}$ denotes the logit of $i$-th class, $\omega_{j} ^{i}$ is the weight of the fully connected layer (FC layer) corresponding to class $i$ for $\mathit{GAP}(F_{j})$. According to \cite{CAM}, we can obtain the CAM corresponding to category $i$ by:
\begin{equation}
\label{mmm}
\begin{aligned}
\mathit{CAM}_{i}(x,y)=\sum_{1\leq j \leq C} \omega_{j}^{i}\times f_{j}(x,y).
\end{aligned}
\end{equation}
According to \Cref{eq_logit_Normal} and \Cref{mmm}, the calculation of $L_{i}$ can be written in another form:
\begin{equation}\label{logit_cam}
\begin{aligned}
L_{i}&=\dfrac{1}{W \times H}\sum_{x,y}\mathit{CAM}_{i}(x,y)\\
&=\mathit{GAP}(\mathit{CAM}_{i}).
\end{aligned}
\end{equation}
\begin{figure}
\centering
\includegraphics[width=8cm]{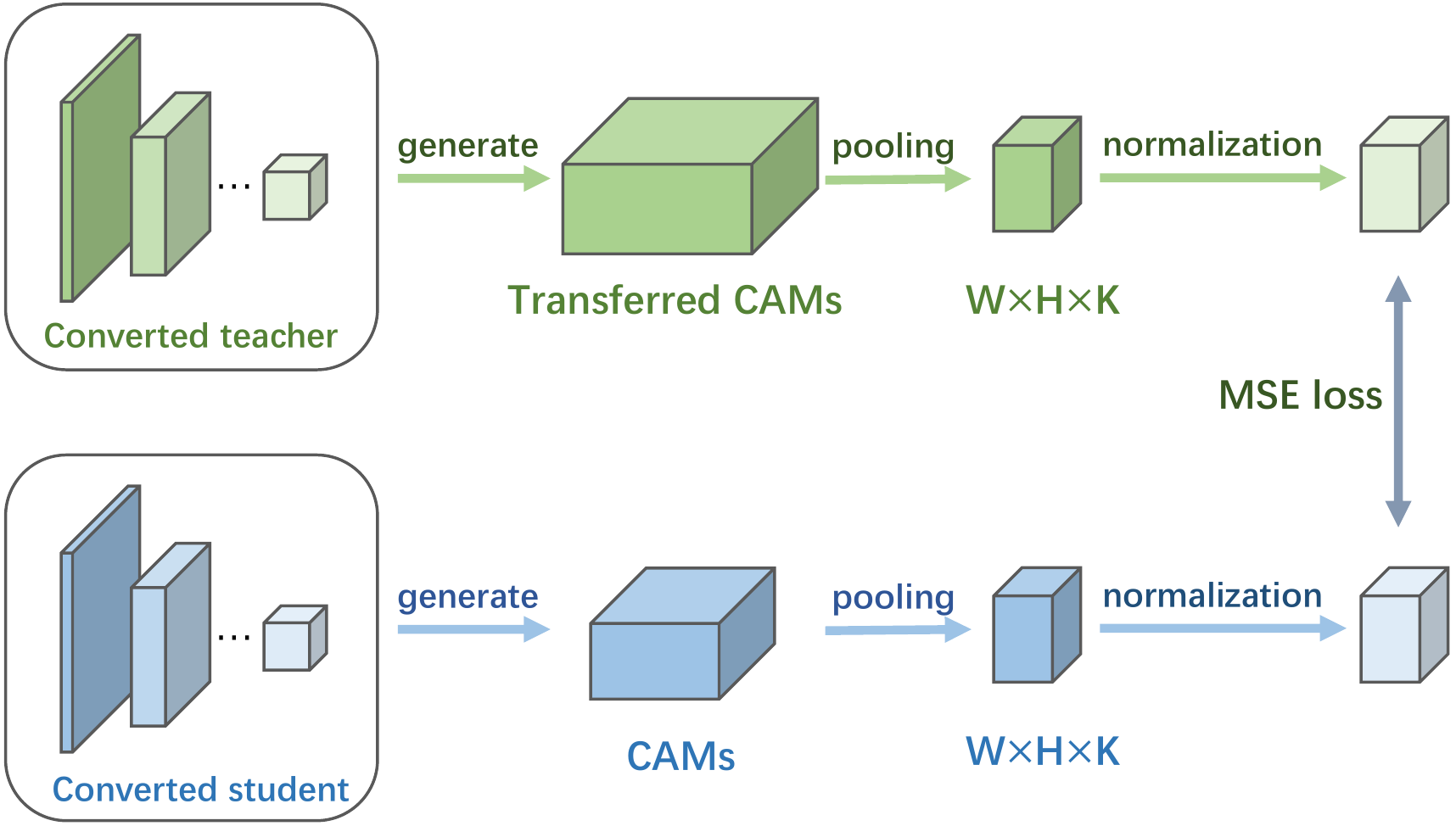}
\caption{Illustration of CAT. During CAT, the structure of teacher and student are converted to our style (\Cref{fig:main}).}
\label{fig:CAT process}
\end{figure}

As reflected in \Cref{logit_cam}, logits can be obtained by computing the average activation of CAMs. Inspired by it, as illustrated in \Cref{fig:main}, we convert the FC layer into a $1\times1$ convolutional layer and move the position of the GAP layer. Then $\Bar{L}_{i}$, the logit of $i$-th class generated by the converted model, can be obtained by: 
\begin{equation}\label{eq_logit_NIN}
\begin{aligned}
\Bar{L}_{i}&=\mathit{GAP}(\mathit{Conv}_{i}(\mathbf{F}))\\
&=\dfrac{1}{W \times H}\sum_{x,y} (\sum_{1\leq j \leq C} \omega _{j} ^{i} \times f_{j}(x,y))\\
&=\mathit{GAP}(\mathit{CAM}_{i}),
\end{aligned}
\end{equation}
where $\mathit{Conv}_{i}$ denotes the converted $1\times1$ convolution kernel that used to separate features corresponding to $i$-th class from $\mathbf{F}$, and $\omega_{j} ^{i}$ is its weight of $j$ channel. As reflected in Eqn\eqref{logit_cam} and Eqn\eqref{eq_logit_NIN}, the conversion does not change the value of its prediction score (i.e., logits). And class activation maps can be obtained during the classification of the converted model. 

As reflected in Eqn\eqref{eq_logit_NIN}, the classification process of the converted model can be viewed in two steps: (1) the model exploits its capacity to identify class discriminative regions of input and generate CAMs, (2) the model outputs prediction score of each category by computing the average activation of the corresponding CAM. Considering that the model makes predictions by simply comparing the average activation of CAMs, possessing the capacity to identify class discriminative regions of input is critical for CNN to perform classification. To examine if this capacity can be obtained and enhanced by offering hints indicating class discriminative regions of input to the trained model, we propose class attention transfer.
%-------------------------------------------------------------------------

% \begin{figure}
% \centering

% \includegraphics[width=8.5cm]{images/2.pdf}
% \caption{Illustration of the process of CAT. During CAT, both CAMs generated by the model being trained and the transferred CMAs provided by the producer are pooled in size, where plane size W$\times$H is defined before the training and K is the number of target classes. After the normalization, the loss of CAT is obtained by computing MSE between them.} 
% \label{fig:CAT process}
% \end{figure}

\subsection{Class Attention Transfer}
\label{CAT}

The purpose of CAT is to examine if a model can obtain the capacity to identify class discriminative regions of input by transferring \textbf{only} CAMs. Thus, during CAT, the trained model is not required to perform classification, and any information related to the category of the training set data (e.g., ground-truth labels and logits) is not released to the trained model. In practice, we utilize a pre-trained model with the converted structure to generate the transferred CAMs. The illustration of the process of CAT is shown in \Cref{fig:CAT process}, while the formal description is shown below.
% The purpose of CAT is to train a model to acquire the capacity to identify class discriminative regions (i.e., generate correct class attention maps), which is realized by forcing the model to mimic the transferred class attention maps. During CAT, only normalized CAMs are fed into the model, any information related to the classes of the training set (e.g., hard labels and logits) is not involved. The total loss we used is:

For a given input, let $\mathbf{A}\in\mathbb{R}^{K\times W \times H}$ denotes the CAMs generated by the converted structure, where $K$ is the number of categories contained in the classification task, $W$ and $H$ denote the width and height of the generated CAM respectively. $A_{i}\in\mathbb{R}^{W \times H}$ represents the $i$ channel of $\mathbf{A}$, which is the CAM corresponding to category $i$. And $S$, $T$ denote student and teacher correspondingly. Besides, we use the average pooling function $\phi$ to reduce the resolution of the transferred CAMs, to improve the performance of CAT (\Cref{sec:Further exploration of CAT}). Then CAT's loss function can be defined as:
\begin{equation}\label{eq_only_CAT}
\begin{aligned}
\mathcal{L}_{CAT}=\sum_{1\leq i \leq K}\dfrac{1}{K}\Vert\dfrac{\phi(A_{i}^{T})}{\Vert \phi(A_{i}^{T}) \Vert_2}-\dfrac{\phi(A_{i}^{S})}{\Vert \phi(A_{i}^{S}) \Vert_2}\Vert_2^{2}.
\end{aligned}
\end{equation} 

As can be seen, we perform $l_2$ normalization on $\phi(A_{i}^{T})$ and $\phi(A_{i}^{S})$ ($l_1$ normalization can be used as well), to ensure that information related to the category of input is not released to the trained model during CAT, considering that the average activation of CAM indicates the prediction score (\Cref{logit_cam}). Besides, note that here we transfer CAMs of all categories, which is based on our finding that CAMs of all categories both contain beneficial information for CAT (\Cref{sec:Further exploration of CAT}).

Our core findings through the experiments with CAT are presented as follows, while the corresponding experimental verification and their detailed analysis can be found in \Cref{sec:Further exploration of CAT}.
% \begin{figure}[!ht] 
% \centering
% \begin{minipage}[b]{1\linewidth} 
% \begin{minipage}[b]{0.326\linewidth} 
% \centering
% \includegraphics[width=\linewidth,height=\linewidth]{images/tiger/top1_tiger.jpg}
% Top1 tiger
% \includegraphics[width=\linewidth,height=\linewidth]{images/tiger/top40_cockroach.jpg}
% Top40 cockroach
% \end{minipage}
% \hfill
% \begin{minipage}[b]{0.326\linewidth} 
% \centering
% \includegraphics[width=\linewidth,height=\linewidth]{images/tiger/top2_leopard.jpg}
% Top2 leopard
% \includegraphics[width=\linewidth,height=\linewidth]{images/tiger/top50_mountain.jpg}
% Top50 mountain
% \end{minipage}
% \hfill
% \begin{minipage}[b]{0.326\linewidth} 
% \centering
% \includegraphics[width=\linewidth,height=\linewidth]{images/tiger/top3_fox.jpg}
% Top3 fox
% \includegraphics[width=\linewidth,height=\linewidth]{images/tiger/top60_bottle.jpg}
% Top60 bottle
% \end{minipage}
% \end{minipage}
% \vfill
% \caption{Visualization of CAMs corresponding to categories with Top $n$ prediction score.}
% \label{Fig: tiger}
% \end{figure}

\begin{itemize}
\item The capacity to identify class discriminative regions of input can be obtained and enhanced by transferring CAMs.
\item CAMs of all categories both contain beneficial information for CAT.
\item Transferring smaller CAMs performs better.
\item For CAT, the critical information contained in the transferred CAMs is the spatial location of the regions with high activation in them rather than their specific value.
\end{itemize}

\subsection{CAT-KD}
After validating the effectiveness of CAT, we apply CAT to knowledge distillation and name it CAT-KD. The loss function of CAT-KD is:
\begin{equation}\label{eq_our_total_loss}
\begin{aligned}
\mathcal{L}_{KD}=\mathcal{L}_{CE}+\beta\mathcal{L}_{CAT},
\end{aligned}
\end{equation} 
where $\mathcal{L}_{CE}$ denotes the standard cross-entropy loss, and $\beta$ is the factor used to balance the CE loss and CAT loss.

Different from previous KD methods, we present how the \textit{knowledge} transferred by CAT-KD helps to improve the performance of the student network: by improving its capacity of identifying class discriminative regions. Besides, through experiments with CAT, we analyze and reveal several properties of the $knowledge$ transferred by our method. This further enhances the interpretability of CAT-KD.
%------------------------------------------------------------------------
\section{Experiments}
\label{4}
\subsection{Datasets and Implementation Details}
\label{4.1}
\noindent
\textbf{Datasets.} In the following section we explore CAT and CAT-KD mainly on two image classification datasets: 

(1) CIFAR-100 \cite{krizhevsky2009learning} comprise 32$\times$32 pixel images of 100 categories, the training and validate sets contain 50K and 10K images. 

(2) ImageNet \cite{JiaDeng2009ImageNetAL} is a large-scale dataset for the classification of 1K categories, containing 1.2 million training and 50K validation images.
\\
\\
\noindent
\textbf{Implementation details.} Our implementation for CIFAR-100 and ImageNet strictly follows \cite{reviewKD, DKD}. Specifically, for CIFAR-100, we train all models for 240 epochs with batch size 64 using SGD. The initial learning rate is 0.05 (0.01 for ShuffleNet \cite{shufflenetv1,shufflenetv2} and MobileNet \cite{MobileNet}), divided by 10 at 150, 180, and 210 epochs. For ImageNet, we train models for 100 epochs with batch size 512. The initial learning rate is 0.2 and divided by 10 for every 30 epochs. We experiment with various representative CNN network: VGG \cite{VGG}, ResNet \cite{resnet}, WideResNet \cite{wrn}, MobileNet \cite{MobileNet}, and ShuffleNet \cite{shufflenetv1, shufflenetv2}.

For fairness, all the results of previous methods are either reported in previous papers \cite{reviewKD, DKD} (we keep our training setting the same as theirs) or obtained using codes released by the author with our training setting. All results on CIFAR-100 are the average over 5 trials, while that on ImageNet is the average over 3 trials.

For all experiments reported in \Cref{sec:Further exploration of CAT} and \Cref{CATKD compare}, without special specification, we pool the transferred CAMs into 2$\times$2 during CAT and CAT-KD. More implementation details such as the settings of $\beta$ are attached in the appendix due to the page limits.
\subsection{Exploration of CAT}
\label{sec:Further exploration of CAT}
In this section, we explore several properties of class attention transfer, which not only help to improve the performance and interpretability of CAT-KD but also contributes to a better understanding of CNN. Note that any information related to the category of the training set (e.g., ground-truth labels and logits) is \textbf{not} utilized in the experiments reported in this section.
\begin{figure}
\centering
\includegraphics[width=8.3cm]{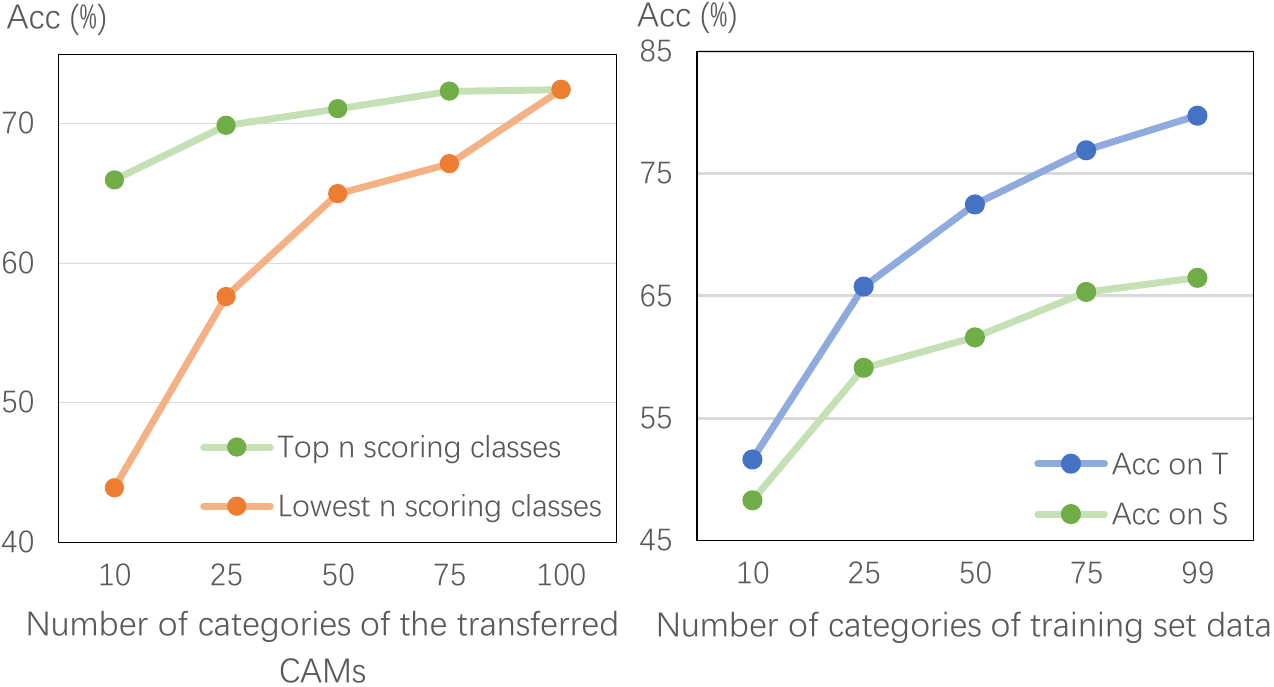}
\vspace{-18pt}
\caption{Accuracy of models trained with CAT on CIFAR-100. \textbf{Left}: Only CAMs of certain categories are transferred, which are selected by two strategies: (1) select categories with top n prediction scores, (2) select categories with the lowest n prediction scores. \textbf{Right}: The training set is reduced to contain data of partial categories only. T: test set of CIFAR-100. S: a subset of T which only contains data of classes that are not contained in the training set.} 
\label{fig:non-target influnce}
\end{figure}
\\
\\
\noindent
\textbf{The capacity of identifying class discriminative regions can be obtained and enhanced by transferring CAMs.} As revealed in \Cref{sec: structure}, being able to identify class discriminative regions of input is critical for CNN to perform classification. Thus, the intensity of this capacity can be evaluated by the model's performance on the classification mission. We perform CAT on ShuffleNetV1, where the transferred CAMs are produced by different models with various accuracy. As the results reported in \Cref{proof for the capacity}, transferring only CAMs can train a model with high accuracy on the classification mission, proving the capacity of identifying class discriminative regions can be obtained by transferring CAMs. Besides, the performance of the trained model is influenced by the accuracy of the model producing the transferred CAMs, indicating that this capacity can be enhanced by transferring more \textit{precise} CAMs.
\begin{figure}[!ht] 
\centering
\begin{minipage}[b]{1\linewidth} 
\hspace{6pt}
\begin{minipage}[b]{0.3\linewidth} 
\centering
\includegraphics[width=\linewidth,height=\linewidth]{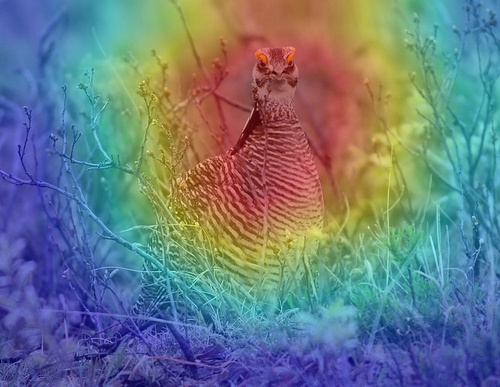}\vspace{0.2pt}
\includegraphics[width=\linewidth,height=\linewidth]{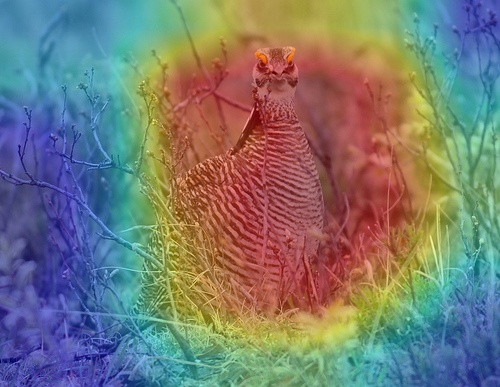}
Top1 prairie fowl
\end{minipage}
\hfill
\begin{minipage}[b]{0.3\linewidth} 
\centering
\includegraphics[width=\linewidth,height=\linewidth]{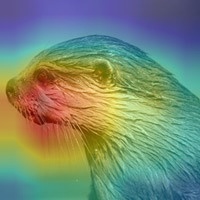}\vspace{0.2pt}
\includegraphics[width=\linewidth,height=\linewidth]{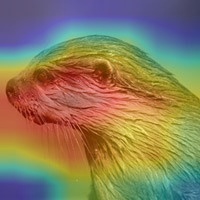}
Top1 otter
\end{minipage}
\hfill
\begin{minipage}[b]{0.3\linewidth} 
\centering
\includegraphics[width=\linewidth,height=\linewidth]{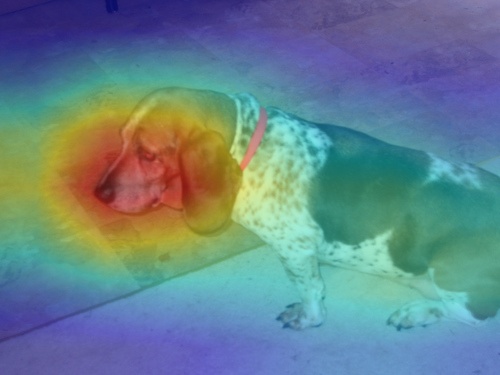}\vspace{0.2pt}
\includegraphics[width=\linewidth,height=\linewidth]{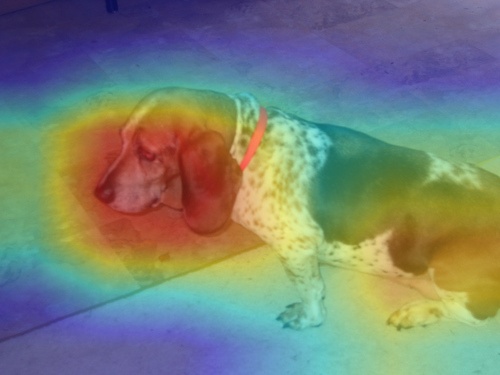}
Top1 basset
\end{minipage}
\hspace{6pt}
\end{minipage}
\vfill
\caption{We set a pre-trained ResNet50 as CAMs producer to train another ResNet50 from scratch with CAT, CAMs are pooled into $2\times2$ during the transfer. The first row shows the visualization of the CAMs generated by the producer, while the CAMs visualized in the second row come from the trained model.}
\label{Fig: comparison between CAT}
\end{figure}
\\
\\
\noindent
\textbf{CAMs of all categories both contain beneficial information for CAT.} For a given input, we can use the method of CAM \cite{CAM} to generate class activation maps for any categories contained in the classification mission. However, though a few non-target categories may share certain similarities (e.g., shape and patterns) with the target category, most of them are completely irrelevant to the input from a human understanding. However, our experiments show that class activation maps of all categories both contain beneficial information for CAT.

We first perform CAT on CIFAR-100 where only CAMs of certain categories are transferred. We designed two strategies to select the categories of the transferred CAMs: (1) select categories with the lowest n prediction scores. (2) select categories with top n prediction scores (the empirical assumption here we make is that the categories with higher prediction scores have more similarities with the target category). As the results reported in \Cref{fig:non-target influnce} (left), while CAMs of classes with higher prediction scores bring more improvement, others are also beneficial for CAT. Besides, we further perform CAT on the reduced CIFAR-100, where CAMs of all classes are transferred but the training set is reduced to contain data of only partial categories. Then the trained model is evaluated on the complete test set and a subset of it which only contains data of classes that are not contained in the training set. As the results reported in \Cref{fig:non-target influnce} (right), interestingly, the trained model achieves high accuracy on the subset, indicating that \textbf{transferring CAMs enables the trained model to classify the categories that are not contained in the training set}. This further proves that non-target CAMs contain beneficial information for CAT even if their categories seem to be irrelevant to the input from a human perspective. 
% \begin{figure}
% \centering
% \includegraphics[width=2.72cm]{images/squirrel/32x4_p8.jpg}
% \hfill
% \includegraphics[width=2.72cm]{images/squirrel/32x4_p4.jpg}
% \hfill
% \includegraphics[width=2.72cm]{images/squirrel/32x4_p2.jpg}
% \includegraphics[width=2.72cm]{images/squirrel/56_p8.jpg}
% \hfill
% \includegraphics[width=2.72cm]{images/squirrel/56_p4.jpg}
% \hfill    
% \includegraphics[width=2.72cm]{images/squirrel/56_p2.jpg}
% \caption{The first row is the visualization of the CAM generated by ResNet32$\times$4 while the second row is that of ResNet110. The size of the CAM in the first column is 8$\times$8, CAMs in the second column are pooled into 4$\times$4, and those in the third column are pooled into 2$\times$2.}
% \label{fig:CAMs_among_different_model}
% \end{figure}
\begin{figure*}[!ht] 
\centering
\begin{minipage}[b]{1\linewidth}
\begin{minipage}[b]{0.3235\linewidth} 
\begin{minipage}[b]{0.3235\linewidth} 
\centering
\includegraphics[width=\linewidth]{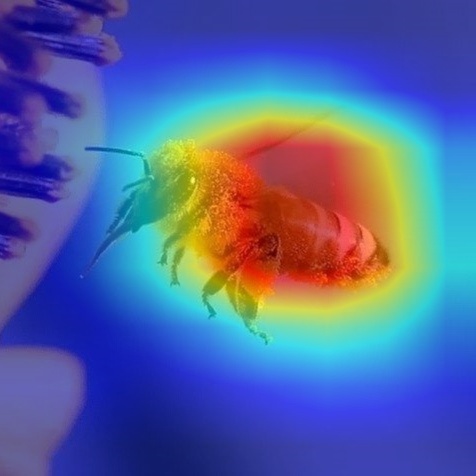}\vspace{1pt}
\includegraphics[width=\linewidth]{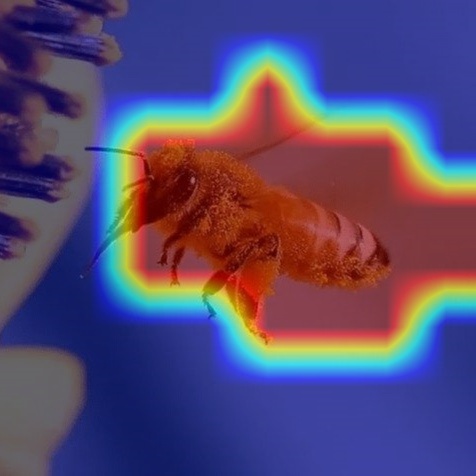}
Top1
\\
bee
\end{minipage}
\hfill
\begin{minipage}[b]{0.3235\linewidth} 
\centering
\includegraphics[width=\linewidth]{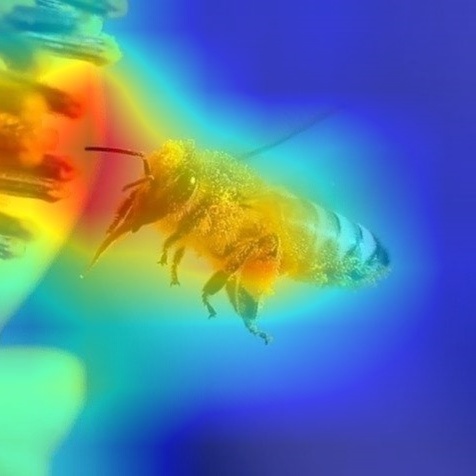}\vspace{1pt}
\includegraphics[width=\linewidth]{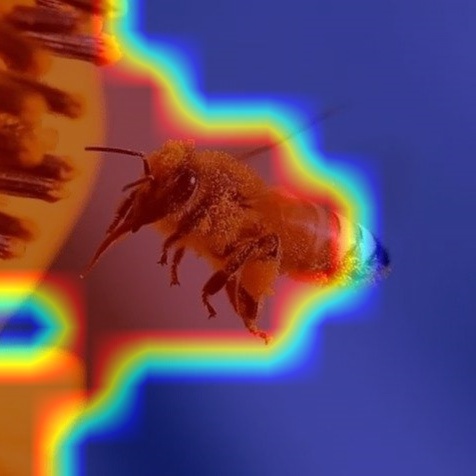}
Top2
\\
sunflower
\end{minipage}
\hfill
\begin{minipage}[b]{0.3235\linewidth} 
\centering
\includegraphics[width=\linewidth]{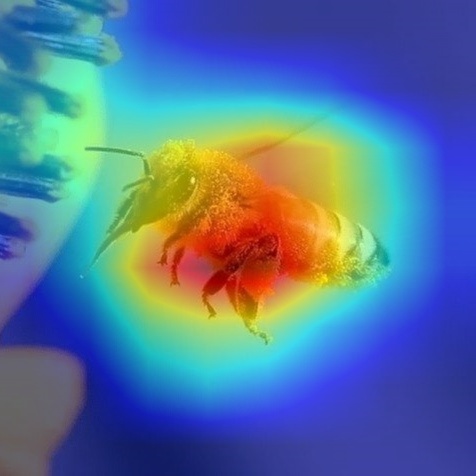}\vspace{1pt}
\includegraphics[width=\linewidth]{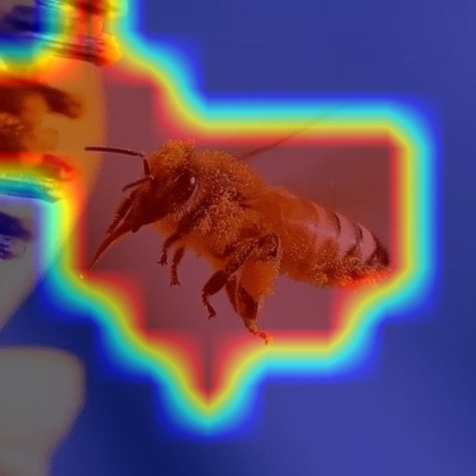}
Top3
\\
tiger
\end{minipage}
\end{minipage}
\hspace{0.5pt}
\hfill
\begin{minipage}[b]{0.3235\linewidth} 
\begin{minipage}[b]{0.3235\linewidth} 
\centering
\includegraphics[width=\linewidth]{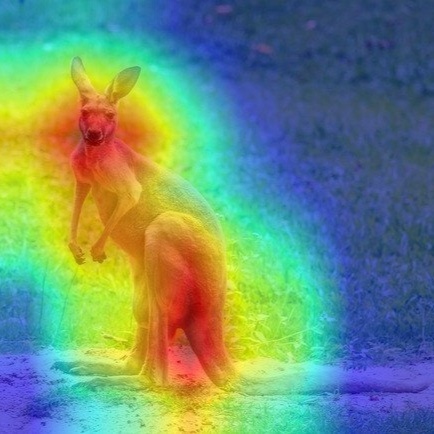}\vspace{1pt}
\includegraphics[width=\linewidth]{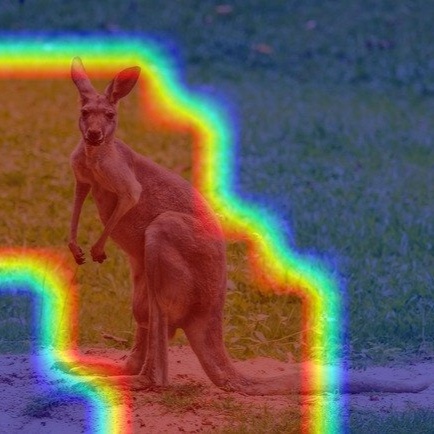}
Top1
\\
kangaroo
\end{minipage}
\hfill
\begin{minipage}[b]{0.3235\linewidth} 
\centering
\includegraphics[width=\linewidth]{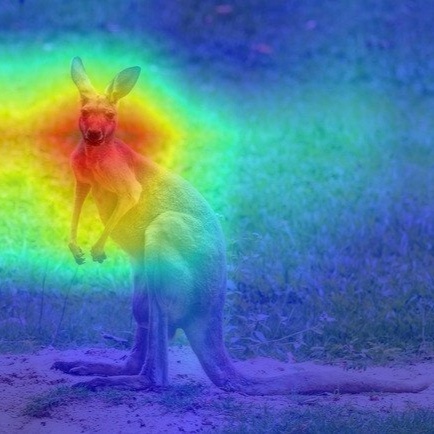}\vspace{1pt}
\includegraphics[width=\linewidth]{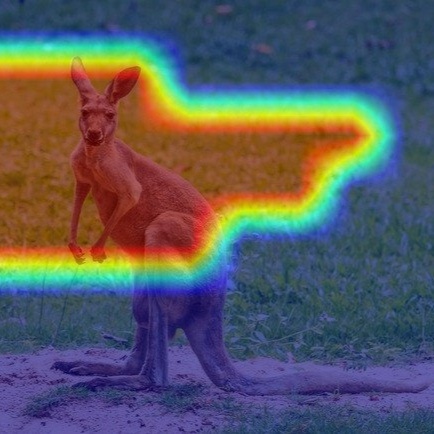}
Top2
\\
rabbit
\end{minipage}
\hfill
\begin{minipage}[b]{0.3235\linewidth} 
\centering
\includegraphics[width=\linewidth]{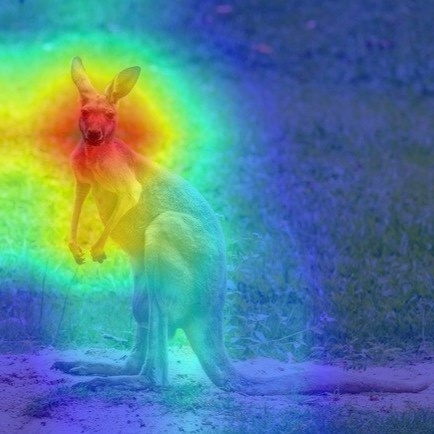}\vspace{1pt}
\includegraphics[width=\linewidth]{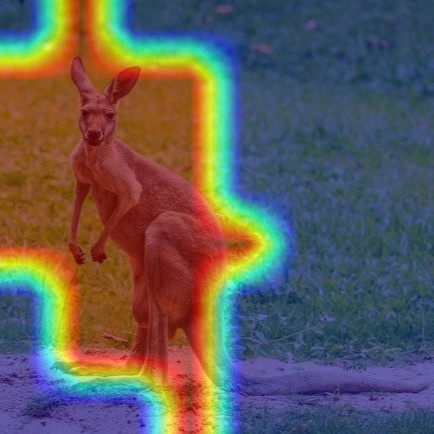}
Top3
\\
fox
\end{minipage}
\end{minipage}
\hspace{0.5pt}
\hfill
\begin{minipage}[b]{0.3235\linewidth} 
\begin{minipage}[b]{0.3235\linewidth} 
\centering
\includegraphics[width=\linewidth]{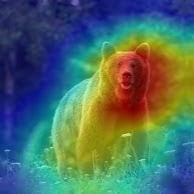}\vspace{1pt}
\includegraphics[width=\linewidth]{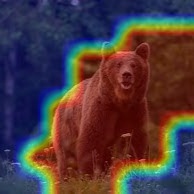}
Top1
\\
bear
\end{minipage}
\hfill
\begin{minipage}[b]{0.3235\linewidth} 
\centering
\includegraphics[width=\linewidth]{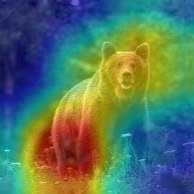}\vspace{1pt}
\includegraphics[width=\linewidth]{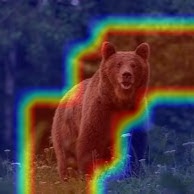}
Top2
\\
cattle
\end{minipage}
\hfill
\begin{minipage}[b]{0.3235\linewidth} 
\centering
\includegraphics[width=\linewidth]{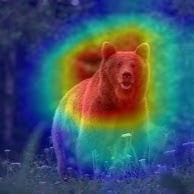}\vspace{1pt}
\includegraphics[width=\linewidth]{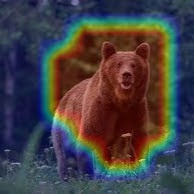}
Top3
\\
otter
\end{minipage}
\end{minipage}
\end{minipage}
\vfill
\vspace{-4pt}
\caption{The first row shows the visualization of the CAMs corresponding to the top 3 predicted categories, while the following row shows the visualization of them after binarization.}
\label{Fig: binary CAMs}
\end{figure*}
\begin{table}[]
\vspace{-4pt}
\centering
\begin{tabular}{c|ccc}
CAM Producer & ResNet56 & ResNet110 & ResNet50 \\
Acc          & 72.34    & 74.31     & 79.34    \\ \hline
CAT          & 72.47    & 74.42     & 76.17   
\end{tabular}
\caption{Accuracy (\%) of ShuffleNetV1 trained with CAT on CIFAR-100. The transferred CAMs are produced by different models with various accuracy.}
\label{proof for the capacity}
\end{table}
\begin{table}[]
\vspace{0pt}
\centering
\resizebox{\columnwidth}{!}{%
\begin{tabular}{ccccc}
\hline
\multirow{2}{*}{Baseline} & Model & ResNet32×4     & ResNet8×4      & ResNet20       \\
                          & Acc   & 79.42          & 72.5           & 69.06          \\ \hline
\multirow{3}{*}{CAT}      & 8×8   & 79.65          & 67.92          & 66.21          \\
                          & 4×4   & \textbf{79.84} & 71.61          & 66.43          \\
                          & 2×2   & 79.71          & \textbf{72.45} & \textbf{66.84} \\ \hline
\end{tabular}%
}
\caption{Accuracy (\%) of various models trained with CAT on the CIFAR-100 test set. During CAT, CAMs are pooled into various sizes. The transferred CAMs are produced by ResNet32$\times$4.}
\label{tab:2}
\end{table}
\\
\\
\noindent
\textbf{Transferring smaller CAMs performs better.} Intuitively, larger CAM contains more detailed hints about the spatial location of the class discriminative regions, then transferring larger CAMs should perform better. However, insufficient accuracy of the model will result in deviations between the highlighted areas in its generated CAM and the actual class discriminative regions of the image (which can be observed in \Cref{fig:CAMs}). Besides, different models differ in their capacity to identify class discriminative regions, which will lead to subtle differences in the generated CAMs. Therefore transferring CAMs with a larger size does not necessarily improve the performance of CAT. Through experiments, we found that performing average pooling on the transferred CAMs, which will expand the highlighted areas of CAMs and reduce the bias between CAMs generated by different models, could alleviate the above issues. As the results reported in \Cref{tab:2}, though pooling blurs the details, transferring smaller CAMs always performs better. Besides, since the pooling operation expands the highlighted areas of CAM, which will make it encompass larger class discriminative regions, transferring pooled CAMs will force the trained model to pay attention to more discriminative regions, which can be observed in \Cref{Fig: comparison between CAT}. In practice, we pool the transferred CAMs into a smaller size to improve the performance of CAT and CAT-KD (normally 2$\times$2).
%-------------------------------------------------------------------------
\begin{table}[]
\vspace{-8pt}
\centering
\resizebox{\columnwidth}{!}{%
\begin{tabular}{cccc}
\hline
\multirow{2}{*}{Baseline} & Model          & ResNet32×4 & ResNet50 \\
                          & Acc            & 79.42      & 79.34    \\ \hline
\multirow{2}{*}{CAT}      & CAMs           & 79.71      & 80.45    \\
                          & Binarized CAMs & 79.35      & 79.65    \\ \hline
\end{tabular}%
}
\vspace{-8pt}
\caption{Results of transferring binarized CAMs. The transferred CAMs are produced by ResNet32$\times$4.}
\label{Tab: transfer binarized CAMs}
\end{table}
\begin{table}[]
\vspace{-5pt}
\centering
\begin{tabular}{cccc}
\hline
Teacher  & \multicolumn{3}{c}{ResNet32×4} \\
Acc      & 77.51    & 79.42    & 81.36    \\ \hline
ReviewKD \cite{reviewKD} & 76.42    & \underline{77.45}    & \underline{77.91}    \\
DKD \cite{DKD}     & \underline{76.58}    & 76.45    & 77.29    \\ \hline
CAT-KD   & 76.36    & 78.26    & 78.84    \\
$\Delta$         & -0.22    & +0.81    & +0.93    \\ \hline
\end{tabular}
\caption{Comparison with two SOTA methods. The student network is ShuffleNetV1. $\Delta$ represents the gap between CAT-KD and the best-performing method among ReviewKD and DKD (marked with underline).}
\label{comparasion with various method}
\end{table}
\begin{table*}[!htb]
\centering
\begin{tabular}{ccccccc}
\hline
\multirow{4}{*}{\begin{tabular}[c]{@{}c@{}}Distillation\\      Mechanism\end{tabular}} &
Teacher &
ResNet32$\times$4 &
WRN40-2 &
ResNet32$\times$4 &
ResNet50 &
VGG13 \\
& Acc      & 79.42 & 75.61 & 79.42 & 79.34 & 74.64          \\ \cline{2-7} 
&
Student &
ShuffleNetV1 &
ShuffleNetV1 &
ShuffleNetV2 &
MobileNetV2 &
MobileNetV2 \\
& Acc      & 70.5  & 70.5  & 71.82 & 64.6  & 64.6           \\ \hline
\multirow{2}{*}{Logits}    & KD \cite{GeoffreyEHinton2015DistillingTK}      & 74.07 & 74.83 & 74.45 & 67.35 & 67.37          \\
& DKD \cite{DKD}     & 76.45 & 76.7  & 77.07 & 70.35 & 69.71          \\ \hline
\multirow{5}{*}{Features}  & CRD \cite{CRD}     & 75.11 & 76.05 & 75.65 & 69.11 & 69.73          \\
& OFD \cite{OFD}     & 75.98 & 75.85 & 76.82 & 69.04 & 69.48          \\
& FitNet \cite{FitNets}  & 73.59 & 73.73 & 73.54 & 63.16 & 64.14          \\
& RKD \cite{RKD}     & 72.28 & 72.21 & 73.21 & 64.43 & 64.52          \\
& ReviewKD \cite{reviewKD} & 77.45 & 77.14 & 77.78 & 69.89 & \textbf{70.37} \\ \hline
\multirow{3}{*}{Attention} & AT \cite{AT}      & 71.73 & 73.32 & 72.73 & 58.58 & 59.4           \\
&
\textbf{CAT-KD} &
\textbf{78.26} &
\textbf{77.35} &
\textbf{78.41} &
\textbf{71.36} &
69.13 \\
& $\uparrow$       &  +6.53  & +4.03   &+5.68  & +12.78 &+9.73          \\ \hline
\end{tabular}
\vspace{-3pt}
\caption{Results on CIFAR-100. Teachers and students have different architectures. $\uparrow$ represents the performance improvement of CAT-KD compared with AT.}
\label{tab: different style}
\end{table*}
\begin{table*}[!htb]
\vspace{-3pt}
\centering
\begin{tabular}{cccccccc}
\hline
& Teacher  & ResNet56       & ResNet110      & ResNet32×4     & WRN-40-2       & WRN-40-2       & VGG13          \\
& Acc      & 72.34          & 74.31          & 79.42          & 75.61          & 75.61          & 74.64          \\ \cline{2-8} 
& Student  & ResNet20       & ResNet32       & ResNet8×4      & WRN-16-2       & WRN-40-1       & VGG8           \\
\multirow{-4}{*}{\begin{tabular}[c]{@{}c@{}}Distillation\\      Mechanism\end{tabular}} &
Acc &
69.06 &
71.14 &
72.5 &
73.26 &
71.98 &
70.36 \\ \hline
& KD \cite{GeoffreyEHinton2015DistillingTK}      & 70.66          & 73.08          & 73.33          & 74.92          & 73.54          & 72.98          \\
\multirow{-2}{*}{Logits}   & DKD \cite{DKD}     & \textbf{71.97} & \textbf{74.11} & 76.32          & \textbf{76.24} & 74.81          & 74.68          \\ \hline
& CRD \cite{CRD}     & 71.16          & 73.48          & 75.51          & 75.48          & 74.14          & 73.94          \\
& OFD \cite{OFD}     & 70.98          & 73.23          & 74.95          & 75.24          & 74.33          & 73.95          \\
& FitNet \cite{FitNets}  & 69.21          & 71.06          & 73.5           & 73.58          & 72.24          & 71.02          \\
& RKD \cite{RKD}     & 69.61          & 71.82          & 71.9           & 73.35          & 72.22          & 71.48          \\
\multirow{-5}{*}{Features} & ReviewKD \cite{reviewKD} & 71.89          & 73.89          & 75.63          & 76.12          & \textbf{75.09} & \textbf{74.84} \\ \hline
& AT \cite{AT}      & 70.55          & 72.31          & 73.44          & 74.08          & 72.77          & 71.43          \\
& \textbf{CAT-KD}     & 71.62          & 73.62          & \textbf{76.91} & 75.6           & 74.82          & 74.65          \\
\multirow{-3}{*}{Attention} &
$\uparrow$&
+1.07 &
+1.31 &
+3.47 &
+1.52 &
+2.05 &
+3.22 \\ \hline
\end{tabular}
\vspace{-3pt}
\caption{Results on CIFAR-100. Teachers and students have the same architecture. $\uparrow$ represents the performance improvement of CAT-KD compared with AT.}
\label{tab: same style}
\end{table*}
\\
\noindent
\textbf{The exact value of the transferred CAMs is not important.} To demonstrate that the role CAMs play in CAT is offering hints about the spatial location of the class discriminative regions of input, we binarize the values of the transferred CAMs to 0 and 1, using their average values as the thresholds. The regions of CAM with values above the threshold are considered as being highlighted, indicating the class discriminative regions of input. Thus, we set the values of these regions to 1 to keep them activated after the binarization. Other regions with values below the threshold are considered unhighlighted, and their values are set to 0. As shown in \Cref{Fig: binary CAMs}, though the specific values of CAMs are lost during the binarization process, the binarized CAMs still contain hints about the spatial location of the class discriminative regions. Note that the threshold can also be specified in other ways (e.g., median). 

As the results reported in \Cref{Tab: transfer binarized CAMs}, although the class discriminative regions obtained by our rudimentary binarization method are not precise, the accuracy of the resulting model dropped by less than one percent, proving that the critical information CAMs contained for CAT is the spatial location of class discriminative regions rather than its exact value. This strongly demonstrates that our method is based on transferring attention.

\subsection{Evaluation of CAT-KD}
\label{CATKD compare}
\begin{figure}
\vspace{-10pt}
\centering
\includegraphics[width=8.2cm]{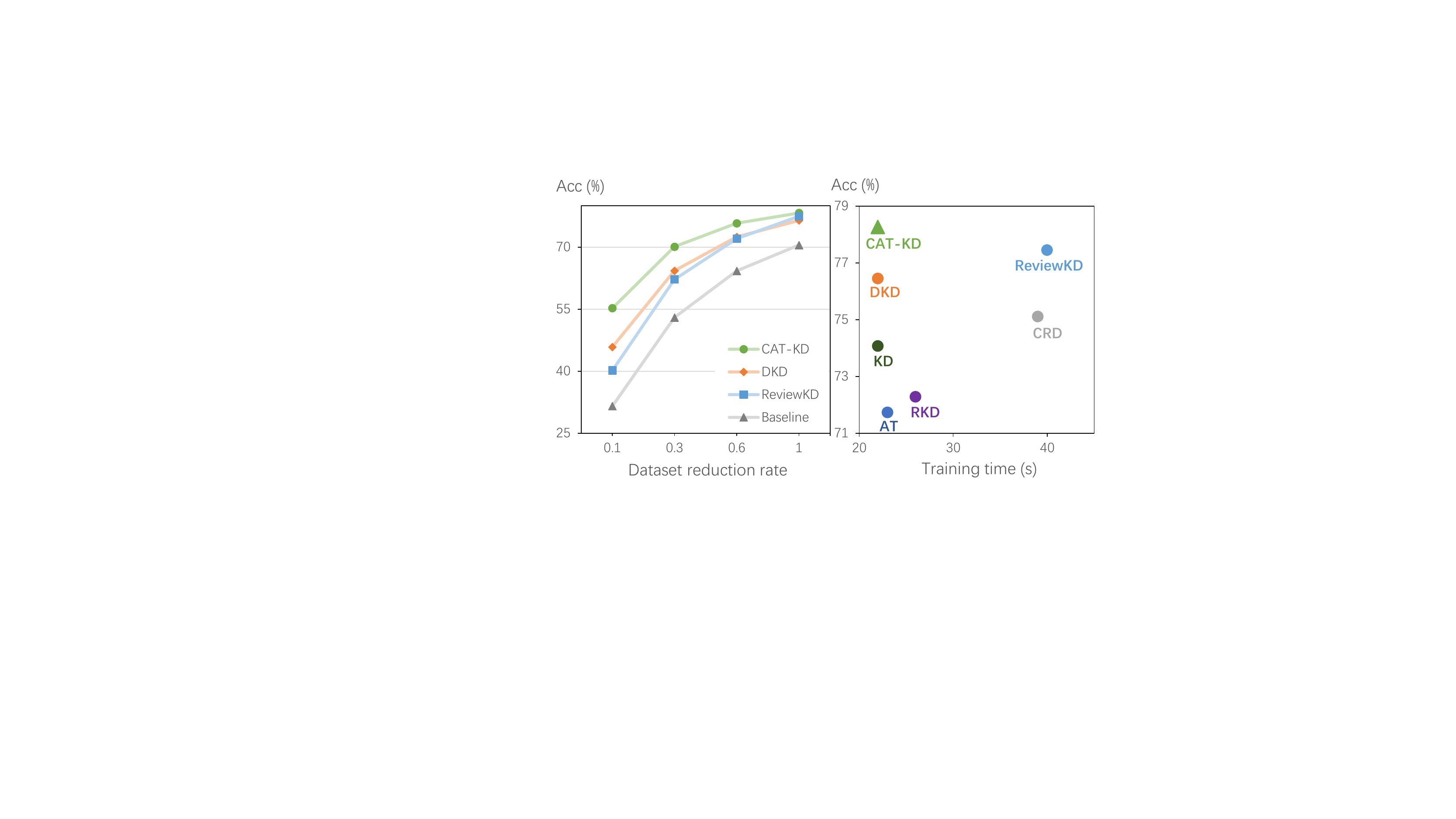}
\vspace{1pt}
\caption{We set ResNet32×4 as the teacher and ShuffleNetV1 as the student. Left: accuracy of students trained with various methods on CIFAR-100, where the training set is reduced at various ratios. Right: comparison of accuracy and training time (per epoch) on CIFAR-100.} 
\label{fig:data reduction}
\end{figure}

Consistent with previous works \cite{CRD,reviewKD,DKD}, we compare the performance of CAT-KD with several representative KD methods. Moreover, we further evaluate our method from two aspects: transferability and efficiency.
\\
\\
\noindent
\textbf{Results on CIFAR-100.} \Cref{tab: different style} reports the results on CIFAR-100 with the teachers and students having different architectures. \Cref{tab: same style} shows the results where teachers and students have architectures of the same style. Notably, our method outperforms the other attention-based method AT \cite{AT} with a large margin (1.07\% $\sim$ 12.78\%). Moreover, CAT-KD achieves comparable or even better performance compared with feature-based distillation method \cite{reviewKD} which requires additional networks and multiple-layer information. Besides, consistent with CAT, the performance of CAT-KD is affected by the accuracy of the teacher: CAMs produced by teachers with lower accuracy contain more incorrect hints about the class discriminative regions of input. To verify this, we further evaluate the impact of the accuracy of the teacher on our method. As the results reported in \Cref{comparasion with various method}, CAT-KD is relatively less effective when the teacher is weak. Thus, as can be observed in \Cref{tab: same style}, the performance of CAT-KD is not the best when the teacher is weak.
\begin{table*}[!htb]
\centering
\renewcommand\arraystretch{1}
\begin{tabular}{ccc|ccc|cc|cc}
& \multicolumn{2}{c|}{} & \multicolumn{3}{c|}{Features} & \multicolumn{2}{c|}{Logits} & \multicolumn{2}{c}{Attention} \\ \hline
& Teacher & Student & OFD \cite{OFD}  & CRD \cite{CRD}  & ReviewKD \cite{reviewKD}      & KD \cite{GeoffreyEHinton2015DistillingTK}   & DKD \cite{DKD}          & AT \cite{AT}   & CAT-KD \\ \hline
Top-1 & 73.31   & 69.75   & 70.81 & 71.17 & \underline{71.61}          & 70.66 & \textbf{71.7} & 70.69 & 71.26  \\
Top-5 & 91.41   & 89.07   & 89.98 & 90.13 & \textbf{90.51} & 89.88 & 90.41         & 90.01 & \underline{90.45} 
\end{tabular}
\vspace{-7pt}
\caption{Results on ImageNet. In this group, we set ResNet34 as the teacher and ResNet18 as the student. The method with the second-best performance is marked with an underline.}
\label{Tab: imagenet1}
\end{table*}
\begin{table*}[!htb]
\vspace{-12pt}
\centering
\renewcommand\arraystretch{1}
\begin{tabular}{ccc|ccc|cc|cc}
& \multicolumn{2}{c|}{} & \multicolumn{3}{c|}{Features} & \multicolumn{2}{c|}{Logits} & \multicolumn{2}{c}{Attention} \\ \hline
& Teacher & Student & OFD \cite{OFD}  & CRD \cite{CRD}  & ReviewKD \cite{reviewKD}      & KD \cite{GeoffreyEHinton2015DistillingTK}   & DKD \cite{DKD}  & AT \cite{AT}   & CAT-KD         \\ \hline
Top-1 & 76.16   & 68.87   & 71.25 & 71.37 & \textbf{72.56} & 68.58 & 72.05 & 69.56 & \underline{72.24}          \\
Top-5 & 92.86   & 88.76   & 90.34 & 90.41 & 91.00          & 88.98 & \underline{91.05} & 89.33 & \textbf{91.13}
\end{tabular}
\vspace{-7pt}
\caption{Results on ImageNet. In this group, we set ResNet50 as the teacher and MobileNet as the student. The method with the second-best performance is marked with an underline.}
\label{Tab2: imagenet2}
\end{table*}
\\
\\
\noindent
\textbf{Results on ImageNet.} \Cref{Tab: imagenet1} and \Cref{Tab2: imagenet2} report the top-1 and top-5 accuracy of image classification on ImageNet. Though the performance of CAT-KD is restricted by the weakness of the teacher network in this setting, our method still outperforms most KD methods.
% \begin{table}[!htbp]
% \centering
% \renewcommand\arraystretch{1}
% \resizebox{\columnwidth}{!}{%
% \begin{tabular}{c|ccc|c}
% Dataset & \multicolumn{3}{c|}{CIFAR-100} & ImageNet \\ \hline
% Model   & VGG13 & ResNet110 & ResNet32$\times$4 & ResNet50 \\
% Acc     & 74.64 & 74.31     & 79.42      & 76.16    \\ \hline
% CAT-KD  & 75.67 & 75.25     & 80.34      & 77.03   
% \end{tabular}%
% }
% \caption{Results on CIFAR-100 and ImageNet where teachers and students are the same. CAMs are pooled into 2$\times$2 during the transfer.}
% \label{Tab: teacher and student are the same}
% \end{table}
% \textbf{Extension} Though the goal of knowledge distillation is to improve the performance of smaller student network by transferring knowledge distilled from bigger teacher network, our CAT-KD breaks through this limitation. Benefiting from performing pooling on transferred CAMs expands the highlighted regions of it, student will be forced to pay attention to more discriminative areas, enabling CAT-KD keeps effective even when teacher and student are the same. As shown in Table \ref{Tab: teacher and student are the same}, CAT-KD achieves consistent improvements on both CIFAR-100 and ImageNet when the students and teachers are the same.
\\
\noindent
\textbf{Transferability.} We perform experiments to compare the transferability of representations to evaluate the generalizability of the \textit{knowledge} transferred by various methods. We use ShuffleNetV1 and MobileNetV2 as the frozen representations extractors, which are either trained from scratch on CIFAR-100 \cite{krizhevsky2009learning} or distilled from ResNet32×4 and ResNet50 with various KD methods. Then linear probing tasks are performed on STL-10 \cite{STL10} and Tiny-ImageNet \cite{JiaDeng2009ImageNetAL} to quantify their transferability. As the results reported in \Cref{Tab:transfer learning}, CAT-KD outperforms other methods by a large margin, indicating the outstanding generalizability of the \textit{knowledge} transferred by our method.
\begin{table}[]
%\vspace{-8pt}
\centering
\begin{tabular}{ccccc}
\hline
Teacher         & \multicolumn{2}{c}{ResNet32×4}   & \multicolumn{2}{c}{ResNet50}    \\
Student         & \multicolumn{2}{c}{ShuffleNetV1} & \multicolumn{2}{c}{MobileNetV2} \\ \hline
Dataset         & STL             & TI             & STL            & TI             \\ \hline
Baseline        & 69.05           & 36.54          & 64.39          & 30.85          \\
KD\cite{GeoffreyEHinton2015DistillingTK}              & 66.61           & 32.56          & 67.81          & 32.37          \\
DKD\cite{DKD}             & 70.73           & 36.77          & 71.05          & 36.48          \\
CRD\cite{CRD}             & 70.68           & 37.85          & 71.46          & 38.75          \\
ReviewKD\cite{reviewKD}        & 71.46           & 38.46          & 66.16          & 32.65          \\
AT\cite{AT}              & 71.36           & 37.36          & 65.1           & 29.13          \\
\textbf{CAT-KD} & \textbf{74.43}  & \textbf{40.73} & \textbf{73.2}  & \textbf{39.87} \\ \hline
\end{tabular}
\caption{Comparison on transferring representations learned from CIFAR-100 to STL-10 (STL) and Tiny-ImageNet (TI).}
\label{Tab:transfer learning}
\end{table}
\\
\\
\noindent
\textbf{Efficiency.} We first compare the performance of multiple KD methods on CIFAR-100, where the training set is reduced at various ratios, to evaluate their dependence on the amount of training data. As the results reported in \Cref{fig:data reduction} (left), CAT-KD is minimally affected by the decrease in the amount of training data, proving the outstanding distillation efficiency of our method. Besides, we further compare the training cost and performance of various KD methods. As reflected in the results reported in \Cref{fig:data reduction} (right), CAT-KD has the highest training efficiency. Since CAT-KD does not require extra parameters, its computational cost is almost the same as logits-based methods. Relatively, feature-based methods require much more computational resources because most of them need additional auxiliary networks to distill features.
% \\
% \\
% \noindent
% \textbf{Limitations and future works.} Due to the performance of CAT being influenced by the accuracy of the model offering CAMs, CAT and CAT-KD are relatively less effective when the teacher (or producer) is weak. However, since the transferred CAMs are \textit{transparent knowledge} and our experiments have shown that modifying transferred CAMs does not influence the effectiveness of CAT (e.g., pooling and binarization), our future work will focus on improving the performance of CAT by modifying the transferred CAMs.
%------------------------------------------------------------------------

\section{Conclusion}
In this paper, we propose CAT-KD which has both high interpretability and competitive performance. More importantly, we demonstrate that the capacity of identifying class discriminative regions of input can be obtained and enhanced by transferring CAMs. Furthermore, we present several interesting properties of transferring CAMs, which contribute to a better understanding of CNN. We hope our findings will help future research on the interpretability of CNN and knowledge distillation.
\\
\\
\noindent
\textbf{Acknowledgement.} We thank the reviewers for their constructive feedback. Part of Hui Li’s work is supported by the National Natural Science Foundation of China (61932015), Shaanxi Innovation Team project (2018TD-007), Higher Education Discipline Innovation 111 project (B16037). Part of Haonan Yan’s work is done when he visits the University of Guelph.

%%%%%%%%% REFERENCES
{\small
\bibliographystyle{ieee_fullname}
\bibliography{egbib}
}

\clearpage

\appendix
\section{Appendix}
\subsection{Cross-entropy loss and CAT loss} 
\label{Why we need both CE loss and CAT loss?}
As we have presented in the paper, the capacity of identifying class discriminative regions is critical for CNN models to perform classification. This capacity can be obtained in two approaches: (1) train models from scratch using cross-entropy loss, and (2) transfer CAMs to the trained model. However, this capacity of the models trained with the first approach is relatively restricted, since during the raw training only hard labels of the training data are offered. For the second approach, though offering hints about the class discriminative regions of input will make it easier for the trained model to obtain this capacity, its performance is also restricted by the accuracy of the model producing the transferred CAMs, because CAMs generated by the model with insufficient accuracy contain incorrect hints for the class discriminative regions of input. 

As the results reported in \Cref{ResNet110} and \Cref{ResNet32x4}, when the CAM producer is stronger than the trained model, only transferring CAMs can let the trained model achieve better performance compared with trained from scratch, since the transferred CAMs are more \textit{correct} than the one that the trained model itself could generate. In contrast, when the CAM producer is weaker than the trained model, transferring CAM is not that effective: its performance is worse than using only the cross-entropy loss function during training. To sum up, (1) compared with the case where only cross-entropy loss function is used, using CAT loss function can further improve the performance of the trained model, (2) using cross-entropy loss function guarantees the performance of the trained model when the CAMs producer is relatively weak. Thus, to ensure the performance of CAT-KD, we need to utilize both cross-entropy loss function and CAT loss function, and balance them correctly.
\subsection{Guidance for balancing CE loss and CAT loss}
As we have discussed in \Cref{Why we need both CE loss and CAT loss?}, properly combining the CAT loss and cross-entropy loss is of great importance for the performance of CAT-KD. As depicted by Eqn \eqref{eq_our_total_loss} in the paper, we use the factor $\beta$ to balance CAT loss and cross-entropy loss. Here we present a guide for tuning $\beta$ from our perspective. As can be observed in \Cref{ResNet110} and \Cref{ResNet32x4}, the transferred CAMs bring more improvement when the teacher is much stronger than the student, while they might not be that beneficial when the capacity of the teacher and student is similar. Thus, the optimal value of $\beta$ should be positively correlated with the capacity of the teacher, but negatively correlated with the capacity of the student. The relevant experimental verification is reported in \Cref{guidance for b}.
% Please add the following required packages to your document preamble:
% \usepackage{multirow}
\begin{table}[]
\centering
\resizebox{\columnwidth}{!}{%
\begin{tabular}{c|ccc}
CAM producer  & ResNet56  & ResNet110 & ResNet32$\times$4 \\
Acc           & 72.34     & 74.31     & 79.42      \\ \hline
Trained Model & ResNet110 & ResNet110 & ResNet110  \\
Acc           & 74.31     & 74.31     & 74.31      \\ \hline
CAT           & 71.86     & 74.54     & 78.13     
\end{tabular}%
}
\caption{Accuracy (\%) of ResNet110 trained with CAT on CIFAR-100 validation set, where the transferred CAMs are produced by various networks.}
\label{ResNet110}
\end{table}
% Please add the following required packages to your document preamble:
% \usepackage{graphicx}
\begin{table}[]
\centering
\resizebox{\columnwidth}{!}{%
\begin{tabular}{c|ccc}
CAM producer  & ResNet56   & ResNet50   & ResNet32$\times$4 \\
Acc           & 72.34      & 79.34      & 79.42      \\ \hline
Trained Model & ResNet32$\times$4 & ResNet32$\times$4 & ResNet32$\times$4 \\
Acc           & 79.42      & 79.42      & 79.42      \\ \hline
CAT           & 72.56      & 78.96      & 79.65     
\end{tabular}%
}
\caption{Accuracy (\%) of ResNet32$\times$4 trained with CAT on CIFAR-100 validation set, where the transferred CAMs are produced by various networks.}
\label{ResNet32x4}
\end{table}
\begin{table}[]
\centering
\begin{tabular}{cc|ccc}
                   & Teacher & ResNet56 & WRN-40-2 & ResNet32×4 \\
                   & Acc     & 72.34    & 75.61    & 79.42      \\ \hline
\multirow{5}{*}{$\beta$} & 10      & 74.08        & 74.96        & 74.57          \\
                   & 50      & \textbf{76.28}        & 76.83    & 76.87          \\
                   & 100     & 75.84        & \textbf{77.31}    & 77.42          \\
                   & 300     & 74.78        & 76.71        & 77.86          \\
                   & 600     & 74.63    & 76.43    & \textbf{78.26}     
\end{tabular}
\caption{Accuracy (\%) of the model trained by CAT-KD on CIFAR-100 with various $\beta$ and different teacher. The student network is ShuffleNetV1.}
\label{guidance for b}
\end{table}
\subsection{Normalization in CAT-KD}
\label{normal}
During CAT, we perform $l_{2}$ normalization on the transferred CAMs to ensure information indicating the category of the input is not released to the trained model. However, this process is not necessary for CAT-KD. As can be observed in \Cref{normal same} and \Cref{normal different}, when the teacher and student have different architecture, performing normalization is beneficial for CAT-KD. However, it will become harmful when the teacher and student have similar architectures. A reasonable explanation is that the \textit{dark knowledge} contained in logits, which will be released to the student model if the normalization is not performed, is relatively more beneficial for the student networks that have similar structure to the teacher. This coincides with the phenomenon that logit-based KD methods perform relatively better when the teacher and student have similar structures, which can be observed in \Cref{tab: different style} and \Cref{tab: same style} reported in the paper. Thus, for CAT-KD, normalization is performed when the student and teacher have different structures, while others are not.
\begin{table}[]
\centering
\begin{tabular}{cccc}
\hline
Teacher & ResNet110 & WRN-40-2 & ResNet32×4 \\
Acc     & 74.31     & 75.61    & 79.42      \\
Student & ResNet32  & WRN-16-2 & ResNet8×4  \\
Acc     & 71.14     & 73.26    & 72.5       \\ \hline
(a)      & \textbf{73.62}     & \textbf{75.6}     & \textbf{76.91}      \\
(b) & 73.45     & 75.46    & 76.29      \\ \hline
\end{tabular}
\caption{Accuracy (\%) of students trained with CAT-KD on CIFAR-100, where students and teachers have similar structure. (a): normalization is performed on the transferred CAMs during CAT-KD. (b): without performing normalization.}
\label{normal same}
\end{table}
% Please add the following required packages to your document preamble:
% \usepackage{graphicx}
\begin{table}[]
\centering
\resizebox{\columnwidth}{!}{%
\begin{tabular}{cccc}
\hline
Teacher & ResNet50    & WRN-40-2     & ResNet32×4   \\
Acc     & 79.34       & 75.61        & 79.42        \\
Student & MobileNetV2 & ShuffleNetV1 & ShuffleNetV1 \\
Acc     & 64.6        & 70.5         & 70.5         \\ \hline
(a)     & 70.86       & 77.24        & 77.78        \\
(b)     & \textbf{71.36}       & \textbf{77.35}        & \textbf{78.26}        \\ \hline
\end{tabular}%
}
\caption{Accuracy (\%) of students trained with CAT-KD on CIFAR-100, where students and teachers have different structure. (a): normalization is performed on the transferred CAMs during CAT-KD. (b): without performing normalization.}
\label{normal different}
\end{table}
\subsection{Extensions}
To facilitate future works related to CAT and CAT-KD, here we offer several extensive experiment results.
\\
\\
\noindent
\textbf{Transfer CAMs generated by other methods.} Following \cite{CAM}, many works propose to generate CAM in other ways \cite{GradCAM, GradCAM++, Score-CAM}. Although these methods always consume much more resources, their generated target class's CAM also correctly highlights the class discriminative regions. To examine if CAT is still effective when the transferred CAMs are generated in these generalized ways, we perform CAT on CIFAR-10 and use GradCAM \cite{GradCAM} to generate the transferred CAMs. The trained model's accuracy is only among 10\%-15\%, indicating transferring GradCAM \cite{GradCAM} barely works. We think this is because CAMs of non-target classes generated by the generalized ways \cite{GradCAM, GradCAM++, Score-CAM} do not contain useful information for CAT, though the visualization of their target class's CAM may look better than that of \cite{CAM}.
\\
\\
\noindent
\textbf{Coefficients in CAT loss.} As we have revealed in \Cref{sec:Further exploration of CAT}, transferring CAMs of categories with higher prediction scores will bring more improvement for the trained model. Then an intuitive idea is that the trained model should focus more on mimicking the CAMs of categories with higher prediction scores. However, through experiments, we find that preferentially transferring CAMs of categories with higher prediction scores brings little benefit for CAT and CAT-KD, while it will increase the complexity and cost of the implementation of our method. Thus, as reported in Eqn (5), we consider transferring CAMs of all categories equally important and give them the same coefficient $1/k$.
\subsection{More implementation details.} For all experiments reported in \Cref{4}, without special specifications, the transferred CAMs are pooled into 2×2 during CAT and CAT-KD. For the experiments reported in \Cref{sec:Further exploration of CAT}, since there do not exist comparisons with other methods, we change the batch size to 128 to accelerate the training, while other settings are the same as those reported in \Cref{4.1}.
\\
\\
\noindent
\textbf{Setup.} All experiments are performed on an Ubuntu 16.04.1 LTS 64-bit server, with one Intel(R) Xeon(R) Silver 4214 CPU, 128GB RAM. For experiments on CIFAR-100, we utilize one RTX 2080 Ti GPU with 11GB dedicated memory. For experiments on ImageNet, we utilize four RTX 2080 Ti GPUs.
\\
\\
\noindent
\textbf{Visualization.} All visualizations presented in the paper are generated by ResNet50, which has 76.16\% Acc on ImageNet.
\\
\\
\noindent
\textbf{CAM's original resolution.} For CIFAR-100, the resolution of CAM generated by all the models involved in this paper is 8×8 except ShuffleNet (4×4), ResNet50 (4×4), VGG (4×4), and
MobileNet (2×2). For ImageNet, their original resolution is 7×7.
\\
\\
\noindent
\textbf{Figure 4.} For the experiment reported in \Cref{fig:non-target influnce} (right), the training set is reduced to only contain data of $n$ categories. The reserved categories are the first n categories in the CIFAR-100 default category order.
\\
\\
\noindent
\textbf{Table 3.} Binarization is performed on the transferred CAMs before they are normalized.
\\
\\
\noindent
\textbf{Table 4.} We employ TrivialAugment \cite{TrivialAugment} to obtain the strong teacher ResNet32$\times$4, which has 81.36\% accuracy on CIFAR-100 validation set. The results of DKD \cite{DKD} and ReviewKD \cite{reviewKD} are obtained using author-released code. For fairness, the hyper-parameters of CAT-KD, DKD, and ReviewKD are not changed with the accuracy of the teachers.
\\
\\
\noindent
\textbf{Table 9.} We first use the code released by DKD \cite{DKD} to obtain student models trained with various distillation methods. For the implementation of linear probing experiments, STL-10 and TinyImageNet share an identical setup. More specifically, we train linear fully connected (FC) layers of models for 40 epochs with batch size 128 using SGD. The initial learning rate is 0.1, divided by 10 at 10, 20, and 30 epochs.
\\
\\
\noindent
\textbf{Figure 7.} For the experiments reported in \Cref{fig:data reduction} (left), the training data of each category is reduced by the same proportion. The reduced data is selected in the CIFAR-100 default order. For the results reported in \Cref{fig:data reduction} (right), we evaluate the training time (per epoch) of various KD methods, where one RTX 2080 Ti GPU with 11GB dedicated memory is used.
\end{document}